\documentclass[twocolumn]{article}

\usepackage{arXiv}
\usepackage[utf8]{inputenc}

\usepackage{authblk}
\usepackage{natbib}
\usepackage{caption}

\usepackage[hyphens]{url}
\usepackage{hyperref}

\usepackage[ruled,vlined]{algorithm2e}
\usepackage{pgfplots}
\pgfplotsset{compat=newest}
\usepgfplotslibrary{groupplots}
\usepgfplotslibrary{dateplot}
\usepackage{bbm}
\usepackage{tikz}
\usepackage{amsthm}
\usepackage{soul}
\usepackage{amsfonts, amsmath, amsthm, amssymb, graphicx, verbatim}
\usepackage{tabularx}
\usepackage{booktabs}
\usepackage{mathtools}
\usepackage{multirow}
\usepackage{subcaption}
\usepackage[short]{optidef}

\definecolor{AccessibleBlue}{HTML}{0072B2}
\definecolor{AccessibleOrangeRed}{HTML}{D55E00}
\definecolor{AccessibleGold}{HTML}{E69F00}

\title{A Framework for Measuring Appropriate Reliance on Set-Valued AI Advice}
\author{\textbf{Ranjan Mishra}}
\author{\textbf{Jakob Schoeffer}}
\affil{University of Groningen, The Netherlands \\ \texttt{\{r.p.mishra,j.j.schoeffer\}@rug.nl}}

\date{}

\begin{document}
\twocolumn[
\maketitle

\sloppy

\begin{abstract}
Appropriate reliance on AI advice has become a central research theme in human-AI collaboration. Existing frameworks have focused exclusively on point predictions as AI advice. However, set-valued AI advice (e.g., discrete sets or continuous intervals) is increasingly being used to communicate uncertainty and improve human decision making. In this paper, we develop the first formal framework for measuring appropriate reliance on set-valued AI advice within the sequential judge-advisor paradigm, spanning both classification and regression tasks. For classification, we first introduce the dimensions that are necessary for evaluating set-valued AI advice. We then define two metrics: \textit{correct reliance rate on AI} and \textit{correct reliance rate on self}, which jointly characterize appropriate reliance in this setting. For regression, we introduce \textit{quantity of AI reliance} and \textit{quality of AI reliance}, which respectively measure whether a decision maker utilized the AI advice and whether their reliance helped them get closer to the ground truth relative to their initial estimate. Through the application of our framework, we demonstrate how these metrics capture important nuances in human-AI collaboration that existing measures overlook.
\end{abstract}
\vspace{2em}
]

\section{Introduction}
\label{sec:introduction}

AI systems are increasingly being integrated into high-stakes decision-making domains such as medicine, law, finance, and public administration~\citep{eckhardt2025survey, raees2026people}.
In these settings, a human decision maker receives AI advice and must decide how much weight to give to it.
The common goal is that the output of this collaborative process surpasses what either the human or the AI could achieve alone, a property known as \textit{human-AI complementarity}~\citep{bansal2021does}.
Achieving complementarity requires that the human decision maker relies appropriately on AI advice: following it when it is correct and resisting it when it is not~\citep{schemmer2023appropriate, lee2004trust}.

Research has shown two systematic failure modes in appropriate reliance.
\emph{Automation bias} leads humans to follow AI advice even when it is incorrect, deferring to the system rather than exercising independent judgment~\citep{mosier2018human}.
\emph{Algorithm aversion} leads humans to dismiss AI advice even when it would improve their decisions~\citep{dietvorst2015algorithm}.
Both failures are well studied but almost exclusively in the context of point predictions, where the AI advice is a single label or value~\citep{raees2026people}.
Moreover, existing frameworks for measuring appropriate reliance are built on this binary correct-or-incorrect characterization of AI advice~\citep{schemmer2023appropriate, cabitza2023ai}.
This design assumption limits the scope of appropriate reliance research and practice to a type of AI advice that is no longer the only or even the dominant form in which AI advice is communicated to humans~\citep{eckhardt2025survey}. 

AI systems increasingly produce advice that expresses uncertainty explicitly~\citep{bhatt2021uncertainty}.
For classification tasks, by outputting prediction sets that enumerate the range of plausible labels, and for regression tasks, by outputting prediction intervals that bound the range of plausible values. 
For example, a clinical decision support system might output a prediction set \{\textit{pneumonia, bronchitis, asthma}\} rather than a single diagnosis, or a tool used for house price estimation might output an interval [\$340\text{K}, \$420\text{K}] rather than a point prediction.
To understand how humans respond to set-valued AI advice, researchers have used the sequential \textit{judge-advisor system} (JAS) framework~\citep{bonaccio2006advice}, in which a human first makes an initial decision and then makes a final decision after receiving the AI advice. 

There is growing empirical evidence that set-valued AI advice can improve human decision making (e.g.,~\citep{cresswell2024conformal, zhang2024evaluating, folgado2025conformal}). 
However, most of these studies focus on accuracy as a metric, which is insufficient to characterize reliance because it can conflate appropriate reliance with undesirable reliance patterns like automation bias and algorithm aversion~\citep{raees2026people}. 
For example, a decision maker can achieve high accuracy by blindly following every AI advice~\citep{raees2026people,schoeffer2025ai}. 
This means that accuracy alone is not sufficient as a metric to characterize reliance behavior.

Other studies using the JAS framework measure reliance directly, but focus on the \emph{quantity} (i.e., how much decision makers moved toward the AI advice) rather than \emph{quality}---whether that adjustment was appropriate and helped them get closer to the ground truth.
Taken together, this motivates the central research question guiding our study:

\begin{description}
    \item[RQ] \textit{What does appropriate reliance mean when AI advice is a set (rather than a point prediction)?}
\end{description} 

To address this question, we develop a framework for measuring appropriate reliance on set-valued AI advice using the JAS framework.
For classification tasks, we derive an exhaustive taxonomy of behavioral reliance patterns and separate them as beneficial, detrimental, or ineffective.
We then introduce two new metrics: \textit{correct reliance rate on AI} ($\text{CRR}_{\text{AI}}$), which measures the proportion of instances in which the human decision maker followed the AI advice and their final decision was correct, and \textit{correct reliance rate on self} ($\text{CRR}_{\text{self}}$), which measures the proportion of instances in which the AI advice was not followed and the decision maker exercised independent judgment to reach the correct answer.
Together, these metrics characterize appropriate reliance in the classification setting.

For regression tasks with interval-valued advice, we build upon the \textit{weight of advice (WoA)} metric~\citep{bonaccio2006advice}---a standard behavioral measure that quantifies the extent to which a decision maker adjusts their final estimate toward the AI advice. 
However, WoA is an insufficient metric for appropriate reliance because it cannot distinguish whether adjusting toward the AI advice was beneficial or not. 
A decision maker can move substantially toward the advice and still end up further from the truth if the advice is poorly calibrated. 
To address this, we introduce two complementary metrics that disentangle behavioral adjustment from decision quality: \textit{quantity of AI reliance} ($\text{AIR}_{\text{quant}}$), which provides a well-behaved measure of behavioral adjustment relative to the advice, and \textit{quality of AI reliance} ($\text{AIR}_{\text{qual}}$), which measures the proportional improvement in absolute error attributable to that adjustment. 
Together, these metrics characterize appropriate reliance by separating \textit{how much} the human followed the advice from \textit{how beneficial} that adjustment was to the final outcome.

Using stylized decision-making tasks, we illustrate that the proposed metrics provide important insight into human-AI collaboration beyond what existing measures capture.
Ultimately, our framework makes a conceptual contribution that provides researchers and practitioners with the diagnostic tools needed to identify specific behavioral pathologies and design systems and interventions that foster appropriate reliance.

\section{Background and Related Work}
\label{sec:related_work}

In this section, we review two areas of research most relevant to our work.
First, we examine how uncertainty is communicated in AI advice and how it shapes human decision making, especially in the context of set-valued advice.
Second, we review existing frameworks for measuring appropriate reliance on AI advice and identify their limitations when applied to sets.

\subsection{Uncertainty in Human-AI Collaboration}
\label{subsec:uncertainty}
Communicating uncertainty in AI advice is increasingly recognized as critical for supporting human decision making~\citep{bhatt2021uncertainty, prabhudesai2023understanding}.
Traditionally, AI advice has taken the form of a single point prediction in the form of a definitive label or value, leaving the human to decide whether to accept or reject it.
Set-valued advice offers a fundamentally different form of support: it offers a range of possible values rather than a single definite value. 

Several methods exist for generating such sets.
In classification, a common baseline is the \textit{top-k} approach, which typically selects $k$ labels with the highest softmax probabilities.
However, more robust methods like \textit{conformal prediction} have gained prominence, primarily because they offer rigorous statistical guarantees~\citep{shafer2008tutorial}.
Unlike fixed-size heuristics, conformal methods produce sets that are guaranteed to contain the true outcome with a user-specified probability (e.g., $90\%$).
The method adapts the set size dynamically based on this probability.
This framework supports both classification~\citep{angelopoulos2023conformal} and regression~\citep{romano2019conformalized} tasks. 

Empirical evidence on how humans respond to set-valued AI advice is growing but shows a mixed picture.
For \textit{classification} tasks, \citet{cresswell2024conformal} conduct a randomized controlled trial comparing conformal prediction sets to fixed-size top-$k$ sets across three tasks, finding that conformal sets improve human accuracy across image classification, sentiment classification, and named-entity recognition tasks, which they attribute to the variable set size communicating model uncertainty. 
\citet{zhang2024evaluating} compare conformal prediction sets to top-1 and top-10 predictions for image labeling, finding that prediction sets improve accuracy for hard out-of-distribution instances but reduce accuracy for easy in-distribution instances due to increased cognitive load from larger sets.
\citet{folgado2025conformal} study set-valued versus single-valued support for ECG interpretation among 62 cardiologists, finding that set-valued support significantly improves diagnostic accuracy in complex cases but has a negative effect for simpler cases. 

For \textit{regression} tasks, studies using the JAS framework have examined how human decision makers respond to set-valued AI advice.
\citet{leffrang2025visualizing} study COVID-19 hospitalization forecasting with 95\% prediction interval plots versus ensemble plots and point forecasts, finding that prediction interval plots increase confidence and advice utilization compared to point forecasts, while ensemble plots decrease advice utilization despite conveying more uncertainty.
Moreover, \citet{holstein2025balancing} conduct two experiments on housing price estimation with prediction intervals, finding that interval width consistently reduces how much decision makers move their final estimate toward the advice (WoA), while framing uncertainty as aleatoric versus epistemic has no significant effect on reliance behavior.

Altogether, these studies establish that human decision makers utilize set-valued AI advice differently from point-based ones. 
However, across these studies, a clear limitation is that they mostly focus on accuracy as a metric, despite accuracy being insufficient for characterizing reliance behavior~\citep{schoeffer2025ai}.
As discussed earlier, accuracy conflates appropriate reliance with undesirable behaviors such as automation bias and algorithm aversion.
Even when prior studies measure reliance directly, they typically focus on the extent to which humans adjust toward the AI advice (e.g., through WoA) rather than on whether that adjustment was appropriate.
Our framework addresses this limitation by explicitly conditioning reliance measures on the ground truth, which allows us to distinguish whether reliance on AI advice was beneficial, detrimental, or ineffective.

\subsection{Metrics of (Appropriate) Reliance}
\label{subsec:appropriate_reliance}

Existing metrics of AI reliance can be categorized into two groups: those that measure the \textit{quantity} of the behavioral adjustment toward the AI advice, and those that measure the \textit{appropriateness} (or \textit{quality}) of that shift relative to the ground truth. 

In prior work, the most common metrics for measuring AI reliance focus exclusively on the magnitude of agreement or adjustment; that is, the quantity of reliance.
For classification, a recent analytical review of 56 empirical studies on human-AI decision making finds that \textit{agreement fraction} (21 studies) and \textit{switch fraction} (15 studies) are the most commonly reported metrics for measuring AI reliance~\citep{raees2026people}. 
For regression, WoA serves as the standard behavioral measure for quantifying the human's adjustment toward AI advice~\citep{bonaccio2006advice, bailey2023meta,kahr2024understanding}.
However, these metrics are ``truth-blind.''
A high switch fraction or a large WoA indicates that the human followed the AI advice, but it cannot distinguish whether this was beneficial (i.e., improving accuracy) or detrimental (i.e., leading the human into an error). 
Hence, relying on quantity metrics alone to measure reliance risks misinterpreting blind deference with successful collaboration.

To resolve this ambiguity, recent frameworks have introduced the concept of appropriate reliance. Appropriate reliance requires that human decision makers follow the AI advice when it is correct and resist such advice when it is not~\citep{lee2004trust, schemmer2023appropriate,schoeffer2025ai}.
\citet{schemmer2023appropriate} define the \textit{relative AI reliance rate} (RAIR) and the \textit{relative self-reliance rate} (RSR) as operationalization of appropriate reliance for point-prediction AI advice.
RAIR measures the proportion of instances where a human appropriately follows correct AI advice, conditioned on instances where the human's prior was wrong and the AI was correct.
RSR measures the proportion of instances where a human appropriately resists incorrect AI advice, conditioned on instances where the human's prior was correct and the AI was wrong.
Together, RAIR and RSR form a two-dimensional characterization that separates automation bias (low RSR) from algorithm aversion (low RAIR). 

\citet{cabitza2023ai} extend the framework of \citet{schemmer2023appropriate} to an 8-tuple decision table enumerating all combinations of a human's initial decision, AI advice, and the human's final decision being correct or not, and measure the degree to which AI advice dominates human judgment.
Further, they introduce odds-ratio measures of automation bias and algorithm aversion. 
These metrics have recently been applied to analyze human-AI performance and reliance behaviors across various domains to gain insights that accuracy alone cannot give (e.g.,~\citep{he2023knowing, he2024err, salimzadeh2024dealing, holstein2025balancing}).

All these frameworks share one common limitation: they assume AI advice is a point prediction (right vs. wrong). 
This assumption breaks down in the context of set-valued advice, where the relevant questions are whether the set covers the ground truth, whether the human enters the set, and whether doing so improved their final decision. 
No existing framework conditions on these jointly. 
Our work addresses this gap through the first framework for appropriate reliance on set-valued AI advice.
The framework introduces novel metrics that capture both the quantity and quality of reliance in classification and regression settings and enables analyses of human-AI collaboration that existing reliance metrics cannot support.

\section{Theoretical Framework}
\label{sec:framework}

Our framework focuses on set-valued AI advice: discrete prediction sets for classification tasks and continuous prediction intervals for regression tasks.
We adopt the \textit{sequential} judge-advisor paradigm~\citep{bonaccio2006advice}, in which a human decision maker first forms an independent judgment before observing AI advice, and then submits a final decision. 
The sequential paradigm is essential for measuring reliance: without a pre-advice decision, it is impossible to distinguish whether the human followed the advice or arrived at the same answer independently \citep{raees2026people}.

We follow established notation in the literature as introduced by \citet{mikhaylova2026we}.
Given a decision space $\mathcal{Y}$, let:
\begin{itemize}
    \item $H \in \mathcal{Y}$ denote the human's \emph{initial decision} (sometimes also referred to as \textit{prior belief});
    \item $F \in \mathcal{Y}$ denote the human's \emph{final decision}, recorded after seeing the AI advice;
    \item $y \in \mathcal{Y}$ denote the \emph{ground truth};
    \item $A \subseteq \mathcal{Y}$ denote the set-valued AI advice.
\end{itemize}
In the (multi-class) classification setting, $\mathcal{Y} = \{1, \ldots, K\}$, $K\in\mathbb{N}$, is a finite label space and $A$ is a \textit{discrete} prediction set. 
In the regression setting, $\mathcal{Y} \subset \mathbb{R}$ and $A = [L, U]$ is a \textit{continuous} prediction interval with $L$ and $U$ being the lower and upper bounds of that interval, respectively. 
For regression, we additionally define $M = (L + U) / 2$ as the midpoint of $[L, U]$, which serves as the AI's implicit point estimate of the most likely outcome, in line with prior work~\citep{holstein2025balancing}.

\subsection{Classification}
\label{subsec:classification}

Unlike prior reliance frameworks (e.g.,~\citep{schemmer2023appropriate, cabitza2023ai}) that condition only on the correctness of the AI's \textit{point prediction}, our framework conditions jointly on whether $H$, $F$, and $y$ lie within the \textit{set} $A$.
We consider the following five binary dimensions to get an exhaustive taxonomy of all possible reliance behaviors.

\begin{enumerate}
    \item $\mathbf{1}[H \in A]$: whether the initial decision lies inside the prediction set;
    \item $\mathbf{1}[H = y]$: whether the initial decision is correct;
    \item $\mathbf{1}[y \in A]$: whether the prediction set covers the ground truth;
    \item $\mathbf{1}[F \in A]$: whether the final decision lies inside the prediction set;
    \item $\mathbf{1}[F = y]$: whether the final decision is correct.
\end{enumerate}
These five binary dimensions yield $2^5 = 32$ total combinations. However, 14 are logically impossible given the following constraints: 

\begin{itemize}
    \item $(H \notin A) \wedge (H = y) \wedge (y \in A$) is impossible: if $(H = y) \wedge (y \in A) \rightarrow H \in A$;
    \item $(H \in A) \wedge (H = y) \wedge (y \notin A)$ is impossible: if $(H = y) \wedge (y \notin A) \rightarrow H \notin A$;
    \item $(F \in A) \wedge (F = y) \wedge (y \notin A)$ is impossible: if $(F = y) \wedge (y \notin A) \rightarrow F \notin A$;
    \item $(F \notin A) \wedge (F = y) \wedge (y \in A)$ is impossible: if $(F = y) \wedge (y \in A) \rightarrow F \in A$.
\end{itemize}

\begin{figure*}[t]
\centering
\includegraphics[width=0.95\textwidth]{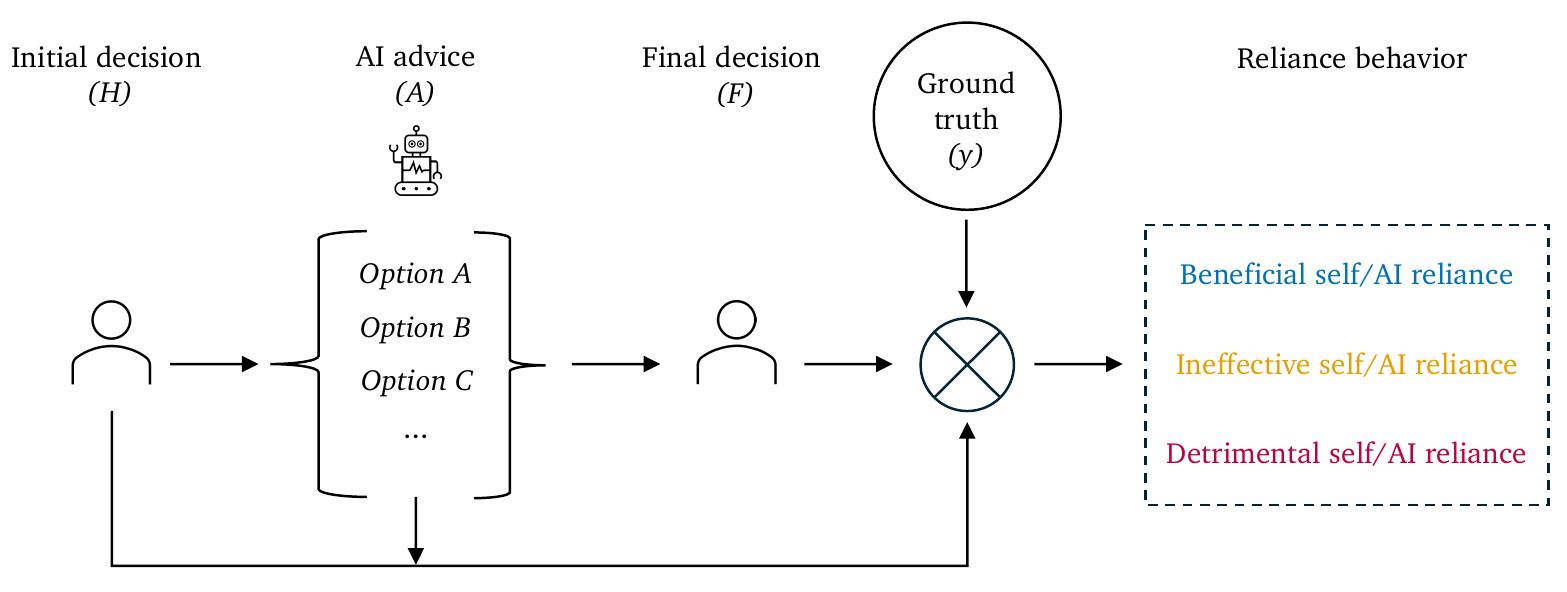}
\caption{The sequential judge-advisor setup for set-valued AI advice in classification. 
A human decision maker first makes an initial decision $H$, then observes the AI advice $A$ (a prediction set), and submits a final decision $F$. 
Comparing $H$, $A$, and $F$ against the ground truth $y$ yields
the reliance behavior, which we classify as beneficial, ineffective, or detrimental and as self- or AI reliance (see Table~\ref{tab:reliance_classification}).}
\label{fig:decision_tree}
\end{figure*}

This yields 18 valid combinations, which we partition into different (instance-level) reliance behaviors.
Those are summarized in Table~\ref{tab:reliance_classification} and
illustrated in Figure~\ref{fig:decision_tree}, together with a descriptive taxonomy of reliance behaviors that is as follows: we call a reliance behavior \textcolor{AccessibleBlue}{\textit{beneficial}} if $F=y$ (i.e., the human's final decision is correct).
We call it \textcolor{purple}{\textit{detrimental}} if $H=y$ and $F \neq y$, and we call it \textcolor{AccessibleGold}{\textit{ineffective}} if $H \neq y$ and $F \neq y$.
Moreover, we denote cases where $F\in A$ as \textit{AI reliance} because a final decision inside the advised set reflects the human following the AI as decision support, whereas cases where $F \notin A$ reflect the human exercising independent judgment, hence \textit{self-reliance}.

\begin{table}[h]
\centering
\setlength{\tabcolsep}{4pt}
\renewcommand{\arraystretch}{1.1}
\small
\resizebox{\columnwidth}{!}{
\begin{tabular}{cc|c|cc|l}
\toprule
\multicolumn{2}{c|}{\textbf{Initial decision}} & \textbf{AI advice} & \multicolumn{2}{c|}{\textbf{Final decision}} & \textbf{Reliance behavior} \\
$H \in A$ & $H = y$ & $y \in A$ & $F \in A$ & $F = y$ & \\
\midrule
0 & 0 & 0 & 0 & 0 & \textcolor{AccessibleGold}{Ineffective self-reliance} \\
0 & 0 & 0 & 0 & 1 & \textcolor{AccessibleBlue}{Beneficial self-reliance} \\
0 & 0 & 0 & 1 & 0 & \textcolor{AccessibleGold}{Ineffective AI reliance} \\
0 & 0 & 1 & 0 & 0 & \textcolor{AccessibleGold}{Ineffective self-reliance} \\
0 & 0 & 1 & 1 & 0 & \textcolor{AccessibleGold}{Ineffective AI reliance} \\
0 & 0 & 1 & 1 & 1 & \textcolor{AccessibleBlue}{Beneficial AI reliance} \\
0 & 1 & 0 & 0 & 0 & \textcolor{purple}{Detrimental self-reliance} \\
0 & 1 & 0 & 0 & 1 & \textcolor{AccessibleBlue}{Beneficial self-reliance} \\
0 & 1 & 0 & 1 & 0 & \textcolor{purple}{Detrimental AI reliance} \\
\midrule
1 & 0 & 0 & 0 & 0 & \textcolor{AccessibleGold}{Ineffective self-reliance} \\
1 & 0 & 0 & 0 & 1 & \textcolor{AccessibleBlue}{Beneficial self-reliance} \\
1 & 0 & 0 & 1 & 0 & \textcolor{AccessibleGold}{Ineffective AI reliance} \\
1 & 0 & 1 & 0 & 0 & \textcolor{AccessibleGold}{Ineffective self-reliance} \\
1 & 0 & 1 & 1 & 0 & \textcolor{AccessibleGold}{Ineffective AI reliance} \\
1 & 0 & 1 & 1 & 1 & \textcolor{AccessibleBlue}{Beneficial AI reliance} \\
1 & 1 & 1 & 0 & 0 & \textcolor{purple}{Detrimental self-reliance} \\
1 & 1 & 1 & 1 & 0 & \textcolor{purple}{Detrimental AI reliance} \\
1 & 1 & 1 & 1 & 1 & \textcolor{AccessibleBlue}{Beneficial AI reliance} \\
\bottomrule
\end{tabular}
}
\vspace{1em}
\caption{The 18 logically valid combinations of the five binary dimensions for classification, each annotated with its interpretable reliance behavior.
A value of 1 means the condition holds and 0 means it does not (e.g., $H \in A = 1$ means the initial decision lies inside the prediction set). Each row is a reliance pattern at the instance level.
A behavior is \textcolor{AccessibleBlue}{beneficial}
if $F = y$, \textcolor{purple}{detrimental} if $H = y$ and $F \neq y$, and \textcolor{AccessibleGold}{ineffective} if $H \neq y$ and $F \neq y$. Cases with $F \in A$ are AI reliance; cases with $F \notin A$ are self-reliance.}
\label{tab:reliance_classification}
\end{table}

Next, we introduce two aggregate metrics that together constitute the \emph{appropriateness of reliance for classification} (AoR-C), building on the conceptual structure from \citet{schemmer2023appropriate}.

\subsubsection{Correct Reliance Rate on AI ($\text{CRR}_{\text{AI}}$)}

We define $\text{CRR}_{\text{AI}}$ over all instances where the AI advice was informative ($y\in A$).\footnote{$\text{CRR}_{\text{AI}}$ is not defined when AI advice is \textit{never} informative. Those cases have no practical relevance when considering a sufficiently large number of instances.}
Among these instances, we measure the proportion of cases where the human decision maker followed the AI advice and their final decision was correct ($F = y$), regardless of their initial decision $H$. 
Mathematically:
\begin{equation*}
\text{CRR}_{\text{AI}} = \frac{N_{y \in A,\, F=y}}{N_{y \in A}}
\in [0,1],\, N_{y \in A}\neq0,
\label{eq:CRR-A}
\end{equation*}
with $N_{y \in A,\, F=y}$ denoting the number of instances classified as \textit{beneficial AI reliance} in Table \ref{tab:reliance_classification}.
Similarly, $N_{y \in A}$ indicates the number of instances where the AI advice was informative; that is, all cases where $y \in A$ in Table~\ref{tab:reliance_classification}.
A value of $\text{CRR}_{\text{AI}}=1$ indicates that whenever the AI advice was informative, the human decision maker followed it, and their final decision was always correct.
A value of $\text{CRR}_{\text{AI}}=0$ indicates that the human decision maker's final decision was never correct on any of the instances where the AI advice was informative, regardless of whether they followed the advice or not.

\subsubsection{Correct Reliance Rate on Self ($\text{CRR}_{\text{self}}$)}

$\text{CRR}_{\text{self}}$ is defined over all instances where the AI advice was uninformative ($y \notin A$).\footnote{$\text{CRR}_{\text{self}}$ is not defined when AI advice is \textit{always} informative. Given a sufficiently large number of instances, this would, in practice, mean that AI advice trivially includes (almost) all possible outcomes.}
Among these instances, we measure the proportion where the human did not follow the AI advice and their final decision was still correct ($F = y$). 
Mathematically: 

\begin{equation*}
\text{CRR}_{\text{self}} = \frac{N_{y \notin A,\, F=y}}{N_{y \notin A}} \in [0,1],\, N_{y \notin A}\neq0,
\label{eq:crrr_self}
\end{equation*}
following the same notation as previously.

A value of $\text{CRR}_{\text{self}}=1$ indicates that whenever the AI advice was uninformative, the human always made the correct final decision using their own judgment, while a value of $\text{CRR}_{\text{self}}=0$ indicates that the human never made the correct final decision when the AI advice was uninformative.

\subsubsection{Appropriateness of Reliance for Classification (AoR-C)}

We characterize the appropriateness of reliance jointly through the two metrics defined above:

\begin{equation*}
\text{AoR-C} = (\text{CRR}_{\text{AI}},\ \text{CRR}_{\text{self}}).
\label{eq:aor}
\end{equation*}
$\text{CRR}_{\text{AI}}$ captures how well the human performs when the AI advice is informative, while $\text{CRR}_{\text{self}}$ captures how well the human uses their own judgment when the AI advice is not informative. 
Together, they provide a two-dimensional characterization of reliance behavior that neither metric offers alone. 
The ideal point is $(1, 1)$, indicating that the human always benefits from the AI advice when it is informative and always relies on their own judgment correctly when it is not.

When either metric is low, the taxonomy in Table~\ref{tab:reliance_classification} allows the failure mode to be diagnosed more precisely.
A low $\text{CRR}_{\text{self}}$ score implies that in the instances where the AI advice was uninformative ($y \notin A$), the human decision maker reached an incorrect final decision ($F \neq y$):
\begin{equation*} 
1 - \text{CRR}_{\text{self}} = \frac{N_{y \notin A, F \neq y}}{N_{y \notin A}}. 
\label{eq:crr_self_diag}
\end{equation*}
The numerator in this equation breaks down into two distinct cases: $N_{y \notin A, F \neq y, F \in A}$ and $N_{y \notin A, F \neq y, F \notin A}$.
If the majority of instances fall into the former, the dominant failure mode is \textit{automation bias}, because the human erroneously relied on uninformative AI advice.
The latter represents a form of \textit{miscalibration} where the human correctly resisted the poor AI advice but was unable to reach the ground truth through independent judgment.

Similarly, a low $\text{CRR}_{\text{AI}}$ score implies that in the majority of instances where the AI advice was informative ($y \in A$), the human failed to reach a correct final decision:
\begin{equation*} 
1 - \text{CRR}_{\text{AI}} = \frac{N_{y \in A, F \neq y}}{N_{y \in A}}. 
\end{equation*}
Two distinct failure modes underlie this metric: $N_{y \in A, F \neq y, F \notin A}$ and $N_{y \in A, F \neq y, F \in A}$. The former is a case of \textit{algorithm aversion}, where the human did not engage with informative AI advice.
The latter indicates another form of \textit{miscalibration}, where the human utilized the AI advice but was unable to identify the correct answer within it---potentially due to the set being too large.
By disentangling these failure modes, our framework provides a taxonomy that functions as a diagnostic tool to design systems and interventions for appropriate reliance.

\subsubsection{Relationship to Existing Metrics}

$\text{CRR}_{\text{AI}}$ and $\text{CRR}_{\text{self}}$ can also be interpreted as a decomposition of decision-making accuracy. 
Weighting each by the proportion of informative and uninformative instances, respectively, recovers accuracy:
\begin{align}
&\mathrm{CRR}_{\mathrm{AI}} \cdot P(y \in A) + \mathrm{CRR}_{\mathrm{self}} \cdot P(y \notin A) \notag \\
& = \frac{N_{y \in A,\, F=y}}{N_{y \in A}} \cdot \frac{N_{y \in A}}{N}
       + \frac{N_{y \notin A,\, F=y}}{N_{y \notin A}} \cdot \frac{N_{y \notin A}}{N} \notag \\
& = \frac{N_{y \in A,\, F=y} + N_{y \notin A,\, F=y}}{N} = \frac{N_{F=y}}{N},
\label{eq:acc_decomposition}
\end{align}
where $N$ without subscript denotes the total number of instances.
Our framework partitions accuracy into two interpretable components and distinguishes reliance patterns that overall accuracy cannot capture.
For example, high overall accuracy may stem either from automation bias or from algorithm aversion~\citep{schoeffer2025ai}; we will exemplify this in Section~\ref{sec:application}.

Compared to the RAIR and RSR metrics of \citet{schemmer2023appropriate}, which condition on the correctness of a point prediction, our metrics condition on $y \in A$ and $y \notin A$, the natural analog for set-valued AI advice.
Unlike agreement and switch fractions~\citep{raees2026people}, which only measure behavioral compliance, our metrics require $F = y$ to ensure that the reliance quality is always conditioned on correctness.

\subsubsection{Aggregation and Reporting}\label{sec:aggregation-classification}
When reporting the aggregate rates, $\text{CRR}_{\text{AI}}$ and $\text{CRR}_{\text{self}}$ should be accompanied by their denominators $N_{y \in A}$ and $N_{y \notin A}$, because each rate is only as reliable as the number of instances it is computed over.
This is especially important when one denominator is small---for example, high AI coverage leaves few uninformative (i.e., $y\notin A$) instances, so $\text{CRR}_{\text{self}}$ may rest on a handful of cases and should be read with corresponding caution.

Moreover, in small-sample and/or high-stakes settings where individual decisions carry more weight, users may also want to report the distribution of instances across the rows of Table~\ref{tab:reliance_classification} rather than only the aggregate $\text{CRR}$ rates.
Finally, Table~\ref{tab:reliance_classification} also captures more granular patterns: for example, comparing $\mathbf{1}[H \in A]$ with $\mathbf{1}[F \in A]$ reveals whether the decision maker entered, left, stayed inside, or stayed outside the set after seeing the advice, and crossing this with $\mathbf{1}[F = y]$ shows whether a change helped.

\subsection{Regression}
\label{subsec:regression}

For regression tasks, AI advice takes the form of a continuous prediction interval $A = [L, U]$ with midpoint $M$.
Analogous to the classification setting, we characterize appropriate reliance along two dimensions: whether the human decision maker utilized the AI advice, and whether doing so improved their final decision ($F$) relative to their initial decision ($H$). 

\subsubsection{Quantity of AI Reliance ($\text{AIR}_{\text{quant}}$)}

We characterize the utilization of AI advice through a new metric called \textit{quantity of AI reliance} ($\text{AIR}_{\text{quant}}$), which measures the proportional reduction (or increase) in distance to the interval midpoint $M$:
\begin{equation*}
\text{$\text{AIR}_{\text{quant}}$} = 
\begin{cases}
    0 & \text{if } H = F = M, \\ 
    -\infty & \text{if } H = M \text{ and } F \neq M, \\
    \frac{|H - M| - |F - M|}{|H - M|} & \text{otherwise,}
\end{cases}
\label{eq:AIR_quant}
\end{equation*}
with $\text{AIR}_{\text{quant}} \in (-\infty, 1]$, where positive values indicate movement toward $M$, and negative values indicate movement away from it; a value of $\text{AIR}_{\text{quant}}=1$ indicates $F=M$ with $H\neq M$; that is, a non-zero movement to the interval midpoint.
Moreover, a value of $\text{AIR}_{\text{quant}}=0$ denotes that the human decision maker's initial and final decision are exactly the interval midpoint $(H = F = M)$ or that the human's initial and final decision are both the same $(H = F)$.
Finally, $\text{AIR}_{\text{quant}}=-\infty$ when the human decision maker's initial decision is exactly the interval midpoint ($H=M$) but their final decision is different from the initial decision ($F \neq M$). 
Mathematically, this means an infinite relative decrease in distance to the midpoint. 
The two special cases handle division by zero and are included for completeness; in practice, exact equality of $H$ or $F$ with $M$ occurs with zero probability given sufficient numerical precision. 
The metric $\text{AIR}_{\text{quant}}$ measures only the \emph{quantity} of reliance, not its \emph{quality}: a decision maker can move substantially toward (or away from) $M$ and still end up further from the ground truth $y$. 

\subsubsection{Quality of AI Reliance ($\text{AIR}_{\text{qual}}$)}

We measure whether relying on the AI advice improved the human decision maker's final decision $(F)$ compared to their initial decision $(H)$ by introducing a metric called $\text{AIR}_{\text{qual}}$, which measures the proportional improvement in absolute error that can be attributed to AI reliance.
Mathematically:
\begin{equation*}
\text{$\text{AIR}_{\text{qual}}$} = 
\begin{cases}
    0 & \text{if } H = F = y, \\ 
    -\infty & \text{if } H = y \text{ and } F \neq y, \\
    \frac{|H - y| - |F - y|}{|H - y|} & \text{otherwise,}
\end{cases}
\label{eq:AIR_qual}
\end{equation*}
with $\text{AIR}_{\text{qual}} \in (-\infty, 1]$,
where 1 indicates a perfectly correct final decision $F = y$, 0 indicates no change from an already correct initial decision $(H = F = y)$ and $-\infty$ indicates the human moved away from a perfectly correct initial decision $H = y, F \neq y$. 
The same considerations regarding the special cases and their near-zero probability in practice apply as for $\text{AIR}_{\text{quant}}$. 

\subsubsection{Appropriateness of Reliance for Regression (AoR-R)}

We characterize the appropriateness of reliance jointly through the two metrics defined above:
\begin{equation*}
\text{AoR-R} = (\text{$\text{AIR}_{\text{quant}}$},\ \text{$\text{AIR}_{\text{qual}}$}).
\end{equation*}
$\text{AIR}_{\text{quant}}$ captures the degree to which the human moved toward (or away from) the AI advice, while $\text{AIR}_{\text{qual}}$ captures whether that movement was beneficial. 
Together, they provide a two-dimensional characterization of reliance behavior that neither metric offers alone.

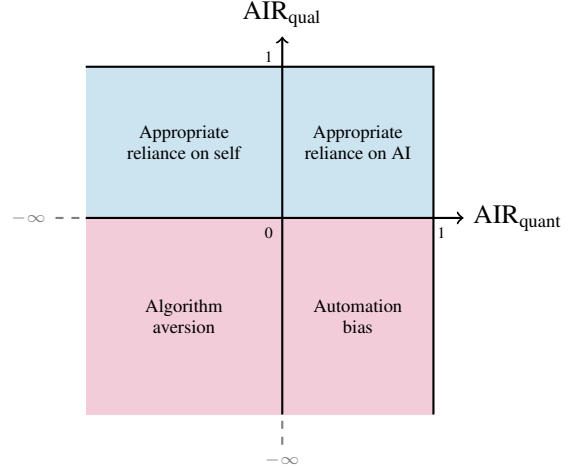
\begin{figure}[ht]
\centering
\begin{tikzpicture}[scale=2]

    \fill[AccessibleBlue!20] (-1.3, 0) rectangle (1, 1);
    \fill[purple!20] (-1.3, -1.3) rectangle (1, 0);

    \draw[dashed, thick, gray] (-1.5, 0) -- (-1.3, 0);
    \draw[->, thick] (-1.3, 0) -- (1.2, 0) node[right] {$\text{AIR}_{\text{quant}}$};
    
    \draw[dashed, thick, gray] (0, -1.5) -- (0, -1.3);
    \draw[->, thick] (0, -1.3) -- (0, 1.2) node[above] {$\text{AIR}_{\text{qual}}$};

    \draw[thick] (1, -1.3) -- (1, 1) -- (-1.3, 1);

    \node[below left, font=\tiny] at (0,0) {0};
    \node[below, font=\tiny, xshift=3pt] at (1,0) {1};
    \node[left, font=\tiny, yshift=4pt] at (0,1) {1};

    \node[left, font=\tiny, gray] at (-1.5, 0) {$-\infty$};
    \node[below, font=\tiny, gray] at (0, -1.5) {$-\infty$};

    \node[align=center, font=\scriptsize] at (-0.65, 0.5) {Appropriate\\reliance on self};
    \node[align=center, font=\scriptsize] at (0.5, 0.5) {Appropriate\\reliance on AI};
    \node[align=center, font=\scriptsize] at (-0.65, -0.65) {Algorithm\\aversion};
    \node[align=center, font=\scriptsize] at (0.5, -0.65) {Automation\\bias};

\end{tikzpicture}
\caption{The AoR-R space for regression, defined by the quantity
($\text{AIR}_{\text{quant}}$) and quality ($\text{AIR}_{\text{qual}}$) of AI reliance. 
The sign of each metric partitions reliance behavior into four
regions: appropriate reliance on AI, appropriate reliance on self, automation bias, and algorithm aversion. Both metrics take values in $(-\infty, 1]$.}
\label{fig:air_quadrants}
\end{figure}

Figure \ref{fig:air_quadrants} distinguishes the distinct reliance behaviors associated with different range of values for $\text{AIR}_{\text{quant}}$ and $\text{AIR}_{\text{qual}}$. 
Positive values for both metrics indicate that the decision maker moved toward the AI midpoint $M$ and improved their final decision accuracy. 
This constitutes \textit{appropriate reliance on AI}. 
When $\text{AIR}_{\text{quant}}$ is negative but 
$\text{AIR}_{\text{qual}}$ is positive, the decision maker moved away from the AI midpoint and used their independent judgment to improve their final decision, indicating \textit{appropriate reliance on self}.
The framework also characterizes two failure modes: the \textit{automation bias} quadrant captures instances where the human adjusted their estimate toward the AI midpoint (positive $\text{AIR}_{\text{quant}}$), but this adjustment increased their final error (negative $\text{AIR}_{\text{qual}}$). 
Negative values for both metrics indicate \textit{algorithm aversion}, meaning that the decision maker moved away from the AI midpoint and, in doing so, worsened their final decision. 
This suggests the AI estimate was actually more accurate than the human's initial decision, but the human rejected the advice to their own detriment. We exemplify these distinct behaviors through examples in Section \ref{sec:application}. 

\subsubsection{Relationship to Existing Metrics}

$\text{AIR}_{\text{quant}}$ is closely related to the \textit{weight of advice} (WoA) metric~\citep{bonaccio2006advice}, a standard measure of advice utilization in the judge-advisor literature. 
WoA measures the signed proportion of the distance between $H$ and $M$ that the human traversed in their final decision $F$; that is, $(F-H)/(M-H).$ 
However, WoA does not account for \textit{overshooting}, which can lead to misleading values.
Consider an example where $H = 12$, $M = 10$, and $F = 4$.
Here, a positive WoA value of $+4$ indicates high reliance but it does not accurately reflect that the human overshot the AI midpoint and landed much farther away from it compared to their initial decision.
Moreover, WoA is not defined for cases where $H = M$.
$\text{AIR}_{\text{quant}}$ introduces important refinements to WoA, namely $(a)$ normalization by absolute distance, and $(b)$ coverage of edge cases, in order to ensure that the metric remains well-behaved and interpretable regardless of the error direction or magnitude.

$\text{AIR}_{\text{qual}}$, on the other hand, is related to the instance-level absolute error change $\Delta\text{AE} := |H - y| - |F - y|$, which is the numerator of $\text{AIR}_{\text{qual}}$.
Specifically, $\text{AIR}_{\text{qual}}$ is the proportional reduction in absolute error---a normalized version of $\Delta\text{AE}$ that adjusts for the initial error $|H - y|$.

\subsubsection{Aggregation and Reporting}
$\text{AIR}_{\text{quant}}$ and $\text{AIR}_{\text{qual}}$ are instance-level measures, which can be aggregated by taking the mean across instances.
We further recommend reporting the proportion of instances falling in each of the four AoR-R quadrants (Figure~\ref{fig:air_quadrants}), which shows how reliance behavior is distributed across appropriate reliance, automation bias, and algorithm aversion. 

Moreover, since $\text{AIR}_{\text{quant}}$ is defined relative to the interval midpoint $M$, it does not directly take into account whether the final decision $F$ falls inside the interval $[L, U]$.
To capture this, users might want to consider the indicator $\mathbf{1}[F \in [L, U]] \in \{0, 1\}$, which equals $1$ when the final decision lies within the AI's prediction interval and $0$ otherwise.
Aggregated over instances, its mean denotes the proportion of final decisions that fall inside the AI advice.
Similar to the classification case (Section~\ref{sec:aggregation-classification}), comparing this against the proportion of instances where $H \in [L, U]$ provides a complementary signal of AI advice utilization---that is, the extent to which the final decisions are drawn towards the AI's recommended range.
We deliberately keep this separate from $\text{AIR}_{\text{quant}}$ rather than folding it into its definition, because conditioning AI reliance on $F \in [L, U]$ would fail to capture cases in which the human adjusts their initial decision in response to the AI advice, without that adjustment resulting in entering or leaving the interval $[L,U]$.

\section{Application}
\label{sec:application}

We illustrate the usefulness of our framework through stylized decision-making scenarios for classification and regression---constructed to demonstrate important analyses that are not possible with existing metrics.

\subsection{Illustration: Classification}
\label{subsec:app_classification}

Consider an occupation prediction task in which an AI system provides
conformal prediction sets over 28 occupations~\citep{de2019bias} as advice.
A human recruiter records an initial assessment $H$, reviews the
prediction set $A$, and submits a final decision~$F$.

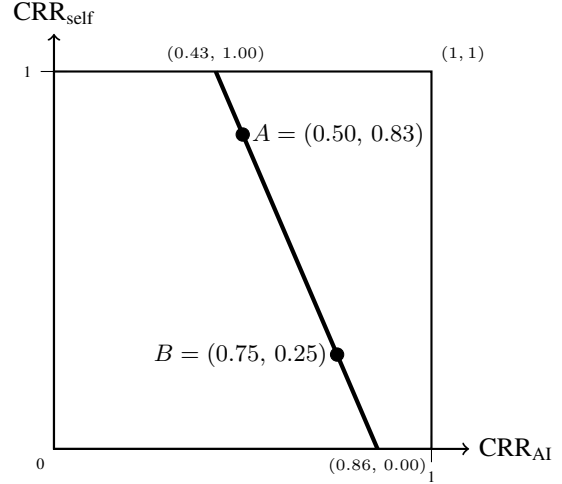
\begin{figure}[ht]
\centering
\begin{tikzpicture}[scale=5]

    \draw[->, thick] (0,0) -- (1.1,0)
        node[right] {$\text{CRR}_{\text{AI}}$};
    \draw[->, thick] (0,0) -- (0,1.1)
        node[above] {$\text{CRR}_{\text{self}}$};

    \node[below left, font=\tiny] at (0,0) {0};
    \draw (1, 1pt) -- (1, -1pt) node[below, font=\tiny] {1};
    \draw (1pt, 1) -- (-1pt, 1) node[left, font=\tiny] {1};
    \node[above right, font=\tiny] at (1,1) {$(1,1)$};

    \draw[thick] (0,0) rectangle (1,1);

    \draw[ultra thick, black] (0.4286, 1.0) -- (0.8571, 0.0);

    \node[above, font=\tiny] at (0.4286, 1.0) {$(0.43,\,1.00)$};
    \node[below, font=\tiny] at (0.8571, 0.0) {$(0.86,\,0.00)$};

    \filldraw[black] (0.50, 0.833) circle (0.5pt)
        node[right, font=\small] {$A=(0.50,\,0.83)$};

    \filldraw[black] (0.75, 0.25) circle (0.5pt)
        node[left, font=\small] {$B=(0.75,\,0.25)$};

\end{tikzpicture}
\caption{Accuracy isoline in the AoR-C space for a final decision-making accuracy of $60\%$ and AI coverage $P(y \in A) = 0.70$. Every
$(\text{CRR}_{\text{AI}}, \text{CRR}_{\text{self}})$ pair on the segment connecting $(0.43, 1.00)$ and $(0.86, 0.00)$ is consistent with this accuracy and coverage. Points $A = (0.50, 0.83)$ and $B = (0.75, 0.25)$ lie on the same isoline but reflect opposing reliance behaviors, which accuracy alone cannot distinguish.}
\label{fig:isoline}
\end{figure}

Suppose the recruiter achieves an overall final decision-making accuracy of 60\% using an AI system with coverage
$P(y \in A) = 0.70$ as advice.
On the surface, a 60\% accuracy in a 28-class task appears satisfactory.
However, without decomposition, we cannot distinguish the underlying reliance behaviors propping up this score.
From the decomposition of Equation~\eqref{eq:acc_decomposition} in Section~\ref{subsec:classification}, we obtain:
\begin{equation*}
    0.60 = 0.70\,\cdot\,\text{CRR}_{\text{AI}}
         + 0.30\,\cdot\,\text{CRR}_{\text{self}},
\end{equation*}
and solving for $\text{CRR}_{\text{self}}$ gives:
\begin{equation*}
    \text{CRR}_{\text{self}} = \frac{0.60 - 0.70\,\cdot\,
    \text{CRR}_{\text{AI}}}{0.30} = 2 - \frac{7}{3}\,\text{CRR}_{\text{AI}}.
\end{equation*}
The feasible segment within $[0,1]^2$ is found by evaluating the
boundary conditions.
Setting $\text{CRR}_{\text{self}} = 1$ gives
$\text{CRR}_{\text{AI}} = 3/7 \approx 0.43$, and setting
$\text{CRR}_{\text{self}} = 0$ gives $\text{CRR}_{\text{AI}} = 6/7
\approx 0.86$; the intersections at $\text{CRR}_{\text{AI}} \in
\{0, 1\}$ fall outside $[0,1]^2$ and are infeasible.
The feasible segment, therefore, connects $(3/7,\,1)$ and $(6/7,\,0)$,
as shown in Figure~\ref{fig:isoline}.

Knowing the accuracy and coverage alone does not identify where
on this segment the recruiter lies---any AoR-C pair on the segment is
consistent with the given accuracy of 60\% and $P(y \in A) = 0.70$.\footnote{For illustration, we are assuming here that the recruiter makes an infinite amount of decisions (i.e., $N\rightarrow\infty)$. In practice, however, $N$ is finite, so the isoline consists of discrete points rather than a continuous line.}
Let us consider two possible points on the isoline, $A$ and $B$, where the recruiter could be. 
If they are at point $A$ where $(\text{CRR}_{\text{AI}},\,\text{CRR}_{\text{self}}) =
(0.50,\,0.83)$, the primary issue is the low $\text{CRR}_{\text{AI}}$ score of $0.50$, which signals a potential failure mode, and our framework allows us to diagnose this further. 
If the majority of the
mistaken decisions with $y \in A$ are such that $F \notin A$ (i.e., the recruiter ignored the informative advice), then the recruiter is exhibiting \textit{algorithm aversion}.
Otherwise, if the majority of the mistaken decisions with $y \in A$ are such that $F \in A$, this means 
the recruiter is engaging with the AI advice but failing to arrive at the correct answer, suggesting a form of miscalibration.

On the other hand, if the recruiter is at point $B$ at $(0.75,\,0.25)$, the main concern here is the low $\text{CRR}_{\text{self}}$ score of $0.25$.
In this case, if the majority of the wrong decisions with
$y \notin A$ are such that $F \in A$,
then the failure mode is \textit{automation bias}, as the human is relying on uninformative AI advice.
Otherwise, if mostly $F \notin A$, then the failure mode might again be a form of miscalibration, where the recruiter correctly resisted the uninformative AI advice but was unable to reach the ground truth through independent judgment. 

Both points $A$ and $B$ satisfy the isoline equation exactly ($0.70 \cdot 0.50
+ 0.30 \cdot 0.83 = 0.70 \cdot 0.75 + 0.30 \cdot 0.25
= 0.60$), yet they represent opposing reliance behaviors.
Behavior in $A$ potentially calls for interventions that help build trust in the AI system; $B$ may, for example, call for training recruiters to critically engage with AI advice.
AoR-C is the only framework that reveals this distinction and helps design for appropriate reliance.

\subsection{Illustration: Regression}
\label{subsec:app_regression}

Consider a house price estimation task in which an AI system provides prediction intervals.
A human appraiser first records an initial estimate $H$, then sees the interval $[L, U]$ with midpoint $M$, and eventually submits a final estimate $F$.
We construct four individual decision-making scenarios depicting different reliance behaviors, shown in Table~\ref{tab:app_regression}, to illustrate the merits of our framework.

\begin{table}[h]
\centering
\resizebox{\columnwidth}{!}{
\begin{tabular}{lcc}
\toprule
& $\text{AIR}_{\text{quant}}$ & $\text{AIR}_{\text{qual}}$ \\
\midrule
Fig.~\ref{fig:regex1}: Algorithm aversion & $-0.05$ & $-0.05$ \\
Fig.~\ref{fig:regex2}: Automation bias  & $0.91$ & $-14.00$ \\
Fig.~\ref{fig:regex3}: Appropriate reliance on AI  & $0.95$ & $0.97$ \\
Fig.~\ref{fig:regex4}: Appropriate reliance on self  & $-0.11$ & $0.67$ \\
\bottomrule
\end{tabular}
}
\vspace{1em}
\caption{Quantity ($\text{AIR}_{\text{quant}}$) and quality
($\text{AIR}_{\text{qual}}$) of AI reliance for the four stylized house price estimation cases (Figure~\ref{fig:regex_all}).
The cases show that behavioral reliance and decision quality can diverge; for example, both automation bias and appropriate reliance on AI have high $\text{AIR}_{\text{quant}}$, but only
the latter has positive $\text{AIR}_{\text{qual}}$.}
\label{tab:app_regression}
\end{table}

\begin{figure*}[t]
\centering

\begin{subfigure}[t]{0.48\textwidth}
  \includegraphics[width=\linewidth]{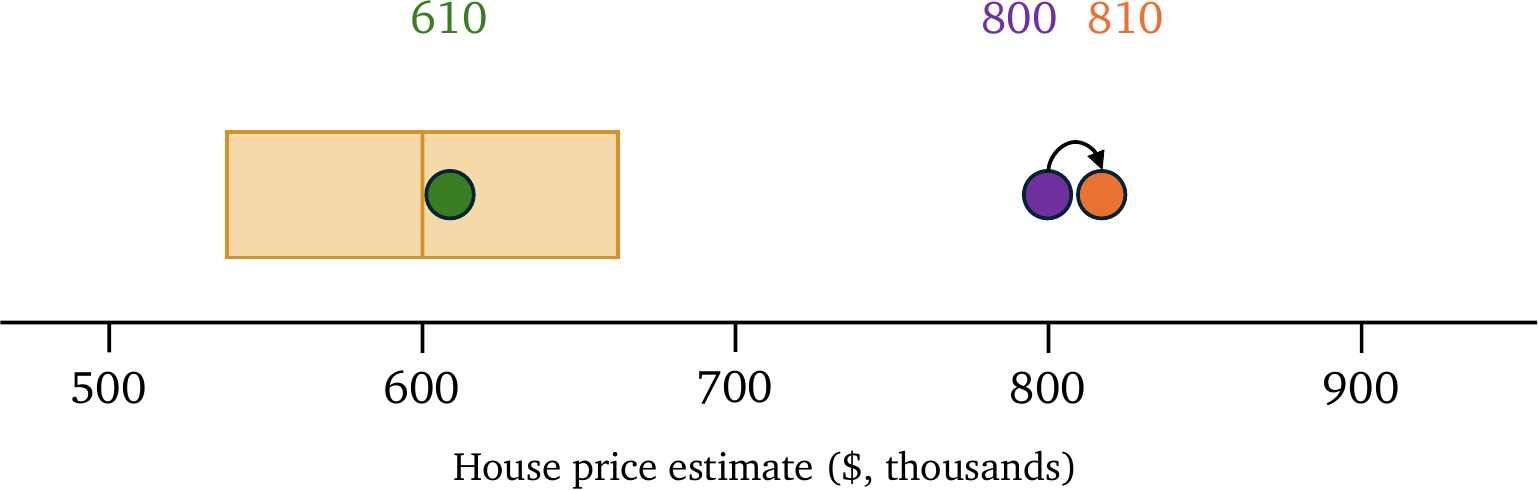}
  \caption{Algorithm aversion}
  \label{fig:regex1}
\end{subfigure}\hfill
\begin{subfigure}[t]{0.48\textwidth}
  \includegraphics[width=\linewidth]{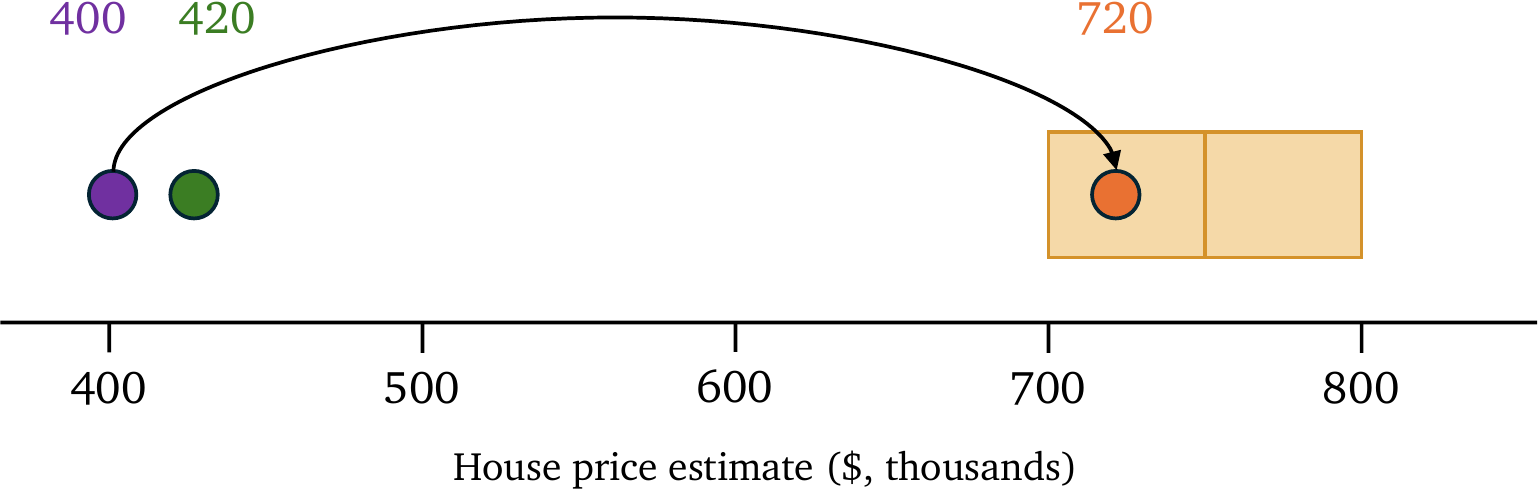}
  \caption{Automation bias}
  \label{fig:regex2}
\end{subfigure}

\vspace{1.5em}

\begin{subfigure}[t]{0.48\textwidth}
  \includegraphics[width=\linewidth]{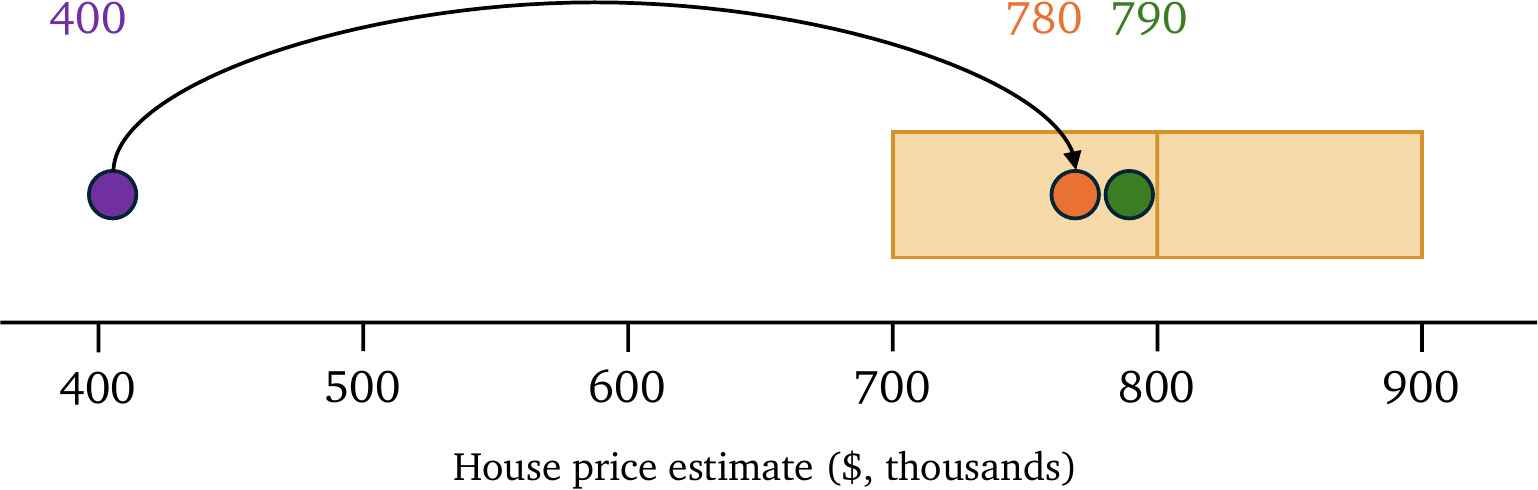}
  \caption{Appropriate reliance on AI}
  \label{fig:regex3}
\end{subfigure}\hfill
\begin{subfigure}[t]{0.48\textwidth}
  \includegraphics[width=\linewidth]{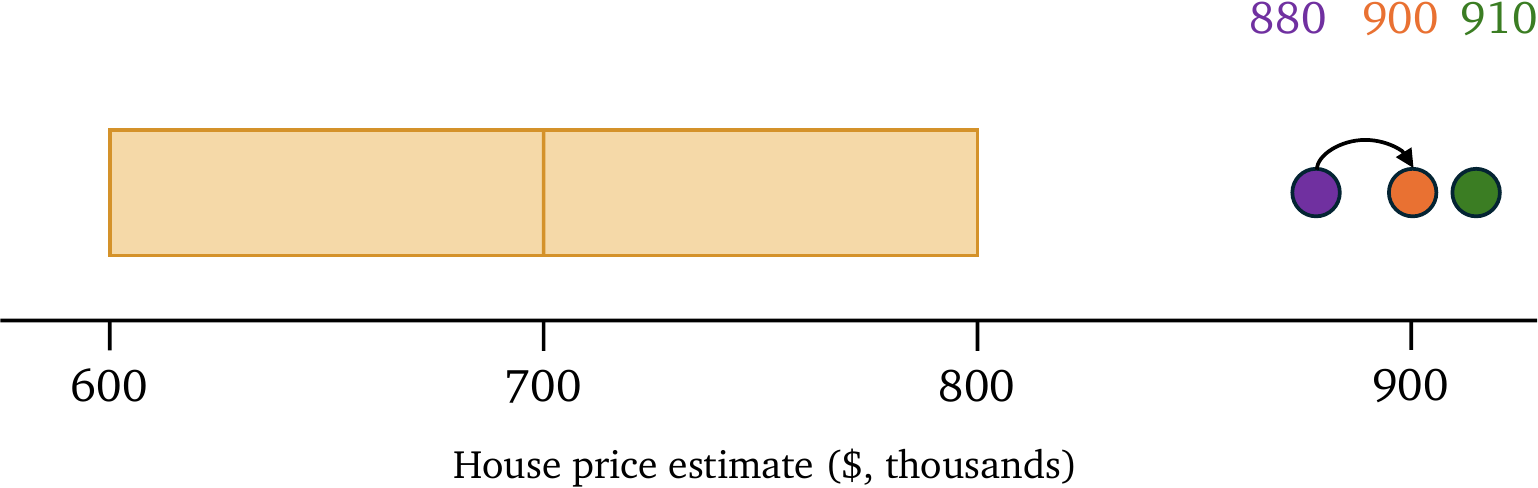}
  \caption{Appropriate reliance on self}
  \label{fig:regex4}
\end{subfigure}

\vspace{1em}

\includegraphics[width=0.5\textwidth]{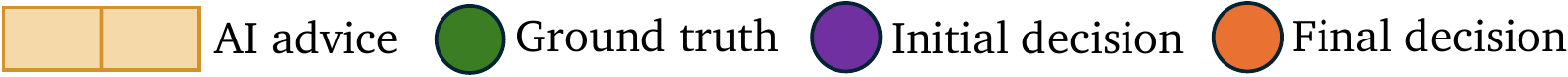}

\caption{Four stylized house-price estimation cases. In each, the appraiser's initial estimate ($H$), the AI interval $[L,U]$ with midpoint $M$, the final estimate ($F$), and the ground truth ($y$) determine the reliance behavior. See Table~\ref{tab:app_regression} for the corresponding $\text{AIR}_{\text{quant}}$ and $\text{AIR}_{\text{qual}}$ values.}
\label{fig:regex_all}
\end{figure*}

\subsubsection{Case: Algorithm aversion}
As visualized in Figure~\ref{fig:regex1}, the appraiser begins with an inaccurate estimate of \$800K while the ground truth is \$610K.
Despite receiving well-calibrated AI advice centered at \$600K, the appraiser moves their final estimate further away from the advice and their initial estimate to \$810K. 
This under-utilization of helpful advice results in a negative quantity of reliance ($\text{AIR}_{\text{quant}} = -0.05$) and a negative quality of reliance ($\text{AIR}_{\text{qual}} = -0.05$), as can be seen in Table~\ref{tab:app_regression}. 
This indicates that the appraiser's independent judgment actively increased the decision error.

\subsubsection{Case: Automation bias}
Figure~\ref{fig:regex2} shows an appraiser with a highly accurate initial estimate of \$400K relative to the \$420K ground truth.
The AI provides a poorly calibrated interval centered at \$750K. 
The appraiser defers strongly to this advice, moving the final estimate to \$720K. 
As shown in Table \ref{tab:app_regression}, while $\text{AIR}_{\text{quant}} = 0.91$ shows significant behavioral reliance, the catastrophic quality score of $\text{AIR}_{\text{qual}} = -14.00$ reveals that this reliance was deeply harmful, increasing the absolute error by 1400\%.

\subsubsection{Case: Appropriate reliance on AI}
As shown in Figure~\ref{fig:regex3}, the appraiser is initially far from the ground truth (\$400K vs. \$790K). 
By recognizing the value of the AI advice, they adjust their final decision to \$780K, very close to the ground truth. 
This adjustment is reflected by high positive scores for both the quantity of reliance ($\text{AIR}_{\text{quant}} = 0.95$) and the resulting quality ($\text{AIR}_{\text{qual}} = 0.97$).

\subsubsection{Case: Appropriate reliance on self}
Finally, Figure~\ref{fig:regex4} shows the appraiser correctly over-ruling a poor AI system.
The appraiser's initial estimate of \$880K is near the ground truth of \$910K, while the AI suggests a much lower interval [\$600K, \$800K] with midpoint \$700K.
The appraiser chooses a final estimate of \$900K, effectively moving away from the AI. The negative quantity of reliance ($\text{AIR}_{\text{quant}} = -0.11$) signals the rejection of advice, while the positive quality ($\text{AIR}_{\text{qual}} = 0.67$) identifies this as appropriate self-reliance that successfully reduced the initial error.

These four cases demonstrate that WoA alone cannot distinguish beneficial from harmful reliance---cases \textit{automation bias} and \textit{appropriate reliance on AI} would both appear as high reliance under WoA, yet one is detrimental and the other appropriate.
Only by jointly reporting $\text{AIR}_{\text{quant}}$ and $\text{AIR}_{\text{qual}}$ can one identify which failure mode is present and design the right intervention.

\section{Discussion}
\label{sec:discussion}

Our framework provides the first formal basis for measuring appropriate reliance on set-valued AI advice.
As AI systems increasingly communicate uncertainty through prediction sets and intervals rather than point predictions alone~\citep{eckhardt2025survey}, existing reliance frameworks based on binary notions of correct versus incorrect advice become insufficient for characterizing human-AI collaboration.
Beyond the metrics themselves, our framework has broader implications for how human-AI collaboration systems are evaluated and designed.

\subsection{Implications for Evaluating Interventions}

A central motivation for measuring appropriate reliance is to support interventions---such as training programs, interface modifications, or explanation strategies---aimed at improving human-AI collaboration.
Prior work has largely evaluated such interventions using downstream decision accuracy or variants of the \textit{weight of advice} (WoA) metric~\citep{bonaccio2006advice}, which are both incomplete characterizations of reliance behavior: accuracy conflates beneficial reliance with undesirable behaviors such as automation bias and algorithm aversion, and WoA lacks a component of decision quality.

Our framework makes the distinction between reliance \textit{quantity} and \textit{quality} actionable.
For example, the isoline analysis in Section~\ref{subsec:app_classification} shows that for a fixed level of accuracy and AI coverage, the AoR-C tuple uniquely locates the system in the reliance space, revealing whether any dominant failure mode is automation bias or algorithm aversion.
This distinction matters because the appropriate intervention for automation bias---reducing blind deference to AI systems---is the opposite of the intervention required for algorithm aversion, which should aim to build calibrated trust in informative advice~\citep{dietvorst2015algorithm,lee2004trust}.
Evaluating interventions using accuracy or WoA alone risks obscuring these differences and may even encourage interventions that improve performance while worsening reliance behavior.

More broadly, our results suggest that evaluations of uncertainty-aware AI systems should move beyond aggregate task performance toward behavioral metrics that explicitly characterize reliance patterns and the quality of human oversight.
This is particularly important in high-stakes domains, where inappropriate reliance on AI systems can lead to substantial downstream harms~\citep{schoeffer2024explanations}.
Recent regulatory frameworks, including the EU AI Act, explicitly require meaningful human oversight for high-risk AI systems to mitigate harms arising from uncritical ``rubber-stamping'' of AI outputs~\citep{laux2025automation,sterz2024quest}.
In such settings, high accuracy alone is insufficient if human decision makers systematically defer to flawed or biased advice.

Our framework also highlights that reliance metrics should be interpreted relative to the performance of the AI system itself.
For AI systems with very high coverage (i.e., $y \in A$) rates, such as conformal predictors that come with strong coverage guarantees \citep{angelopoulos2023conformal}, cases in which humans should override the AI advice become inherently rare.
In such settings, $\text{CRR}_{\text{self}}$ naturally has a small denominator, which reflects that appropriate behavior will often consist of relying on the AI advice.
At the same time, these rare cases where the correct decision lies outside the prediction set become especially important to evaluate.
If humans systematically fail to override the AI in such situations, this may indicate a subtle but consequential form of over-reliance despite otherwise strong aggregate performance.

\subsection{Implications for System Design}

Our framework also has implications that go beyond retrospective evaluation and inform a proactive design of AI-based decision support systems. 
A recurring finding from our application scenarios is that when systems are optimized for accuracy alone, they may inadvertently encourage passive deference rather than genuine engagement with AI advice~\citep{mosier2018human}.
Over time, such patterns risk eroding human critical thinking abilities and contextual knowledge~\citep{gerlich2025ai}.
To mitigate this, systems, including those that issue set-valued advice, should be designed for \textit{verifiability}---the degree to which a human can independently evaluate the correctness of an AI's output---rather than predictive power alone~\citep{fok2024search}.

Our framework provides diagnostic signals needed to identify which specific failure mode a system is exhibiting so that we can design appropriate interventions to address these.
Most existing interventions, especially explainable AI (XAI), are designed for spotting individual point mistakes of AI~\citep{buccinca2021trust, vasconcelos2023explanations,zhang2024you}.
With set-valued advice, the challenge is fundamentally different: because sets are designed to cover the truth most of the time, they almost always contain a mixture of correct and incorrect candidates.
Novel interface designs may be needed---for example, visual cues that communicate the precision-coverage tradeoff or a flag when a set is likely uninformative---to help users differentiate high-signal from low-signal advice~\citep{langer2024effective} and calibrate reliance on sets.

In regression settings, the AoR-R quadrants reveal whether users are struggling with algorithm aversion or automation bias.
For algorithm aversion, interventions that build (calibrated) trust, for example, by showing the model's historical calibration may help. 
For automation bias, skepticism-inducing interventions---such as requiring users to justify a final estimate that deviates substantially from their prior before accepting the AI advice---akin to cognitive forcing~\citep{buccinca2021trust} might help maintain meaningful human oversight rather than passive rubber-stamping~\citep{green2022flaws}.

\subsection{Practical Considerations}
\label{subsec:considerations}
Our framework provides a diagnostic tool for analyzing set-valued advice. Here, we discuss general practical considerations for researchers and practitioners applying the framework.

\paragraph{Inclusion of consensus cases}
In the classification setting (Section \ref{subsec:classification}), the metrics are computed over all instances, including consensus cases where the initial and final decisions both lie in the AI advice set ($H \in A$ and $F \in A$). 
Prior work on reliance for point-prediction AI advice avoids this ambiguity by conditioning only on human-AI disagreement cases ~\citep{schemmer2023appropriate, cabitza2023ai}---this, however, is ill-suited for set-valued advice: as coverage $P(y \in A)$ grows, the initial decision $H$ increasingly falls inside the set, so disagreement becomes rare and coverage-dependent.
Restricting the analysis to those cases would compute the metrics over a shrinking, unrepresentative subset.
We therefore keep consensus instances so the denominators reflect real usage.
Future work could incorporate latent variable modeling to better infer the specific weight of AI influence in such consensus scenarios~\citep{tejeda2022ai}.

\paragraph{Midpoint as reference point in regression}
In the regression setting (Section \ref{subsec:regression}), $\text{AIR}_{\text{quant}}$ is defined relative to the midpoint $M = (L+U)/2$.
This is consistent with WoA~\citep{bonaccio2006advice} and with empirical work that applies WoA to interval-valued advice using a central value~\citep{holstein2025balancing}.
However, the definitions in Section~\ref{subsec:regression} do \textit{not} depend on this choice: $M$ can be replaced by any point estimate the AI reports (e.g., median or mode), which coincides with the midpoint for symmetric intervals but may differ for skewed ones.
Our framework uses the midpoint by default because it needs no distributional assumptions beyond the reported bounds.

\paragraph{Judge-advisor system setup}
Like all studies utilizing the sequential judge-advisor paradigm~\citep{bonaccio2006advice}, our metrics are susceptible to the behavioral artifacts of this setup.
For example, the requirement of an initial decision ($H$) can introduce \textit{anchoring effects}~\citep{tversky1974judgment}, where the human's final decision is tied to their first estimate rather than being a pure reflection of AI reliance---a pattern akin to \textit{egocentric advice discounting}~\citep{yaniv2000advice, yaniv2004receiving}.
At the same time, without collecting $H$, it would not be possible to meaningfully assess how AI advice influences human reliance or changes subsequent decisions~\citep{schemmer2023appropriate}.
This tradeoff is inherent to the paradigm, but its impact can be mitigated through experimental design, for example by allotting deliberation time~\citep{rastogi2022deciding} or \textit{consider-the-opposite} strategies~\citep{adame2016training}.

\section{Conclusion}

Because AI systems increasingly communicate uncertainty through set-valued advice rather than single-point predictions, evaluating whether humans rely on such advice appropriately becomes a central challenge for effective human-AI collaboration.
This work introduces the first formal framework for measuring appropriate reliance on set-valued AI advice across both classification and regression tasks.
For classification, our proposed metrics, $\text{CRR}_{\text{AI}}$ and $\text{CRR}_{\text{self}}$, provide an interpretable decomposition of decision-making accuracy that captures distinct patterns of human reliance and reveals insights that existing metrics cannot detect.
For regression, $\text{AIR}_{\text{quant}}$ and $\text{AIR}_{\text{qual}}$ jointly measure both the quantity and quality of reliance on AI prediction intervals, which addresses a key limitation that prior work has acknowledged but left unresolved.
Together, these metrics provide a principled foundation for understanding reliance behavior in uncertainty-aware AI systems and offer practical tools for evaluating and designing interventions that promote effective human-AI collaboration.

\bibliographystyle{plainnat}
\bibliography{bibliography}

\end{document}